\documentclass[conference]{IEEEtran}

\usepackage{amsmath,amssymb,amsfonts}
\usepackage{graphicx}
\usepackage{float}
\usepackage{textcomp}
\usepackage{xcolor}
\usepackage{cite}
\usepackage{booktabs}
\usepackage{hyperref}
\usepackage{bm}
\hypersetup{hidelinks}

\newcommand{\R}{\mathbb{R}}
\newcommand{\x}{\mathbf{x}}
\newcommand{\z}{\mathbf{z}}
\newcommand{\bs}[1]{\boldsymbol{#1}}

\begin{document}
\raggedbottom

\title{FOCUS: Foot Observation Confidence for Robust Humanoid Proprioceptive Odometry}

\author{%
\IEEEauthorblockN{%
Kaixin Feng\textsuperscript{1},
Angsong Li\textsuperscript{2},
Shaopeng Zhang\textsuperscript{2},
Enyu Li\textsuperscript{2},
Peiwen Lin\textsuperscript{2},
Chuang Wang\textsuperscript{2},\\
You Li\textsuperscript{1},
and Haiyu Lan\textsuperscript{2,}\thanks{$^{\dagger}$Corresponding author: Haiyu Lan.}\textsuperscript{$\dagger$}%
}
\IEEEauthorblockA{%
\textsuperscript{1}Wuhan University\\
\textsuperscript{2}AgiBot%
}%
}

\maketitle

\begin{abstract}
Foot forward kinematics (FK) is widely used to improve proprioceptive legged odometry by providing reliable velocity constraints during foot support. Existing contact-aided estimators generally rely on binary contact decisions to determine whether the FK measurements of an entire foot should be trusted. However, contact does not necessarily imply FK reliability. Dynamic locomotion often involves partial support, toe dragging, and foot slip, causing binary contact decisions to accumulate significant drift over long trajectories. To address this limitation, we propose FOCUS (Foot Observation Confidence from Unannotated Simulation), which predicts a continuous FK reliability weight for each foot instead of estimating binary foot contact. Rather than replacing the model-based estimator, the predicted reliability weights are used to blend FK velocity observations with IMU-propagated body velocity and to adapt the observation covariance of an extended Kalman filter (EKF), enabling smooth reliability-aware fusion without hard contact switching. The network is trained from automatically generated simulation signals using an FK-weighted velocity consistency loss with lightweight simulator-contact regularization, without manually annotated continuous FK-reliability labels. The deployed model relies only on IMU and joint kinematic measurements, making it suitable for hardware platforms with unreliable torque sensing. Experiments demonstrate that FOCUS reduces absolute trajectory error (ATE) by 83.7\% on simulated walking episodes, preserves simulated dynamic-motion fidelity in motion scale and spectral energy, reduces ATE by 70.8\% across 19 real walking segments, and reduces mean ATE by 42.7\% across four real dynamic-motion routines.
\end{abstract}


\section{Introduction}
\label{sec:intro}

Reliable proprioceptive odometry is essential for legged robots operating without external localization. A widely used approach is to fuse inertial measurements with foot forward kinematics (FK) in an extended Kalman filter (EKF)~\cite{bloesch2013state,rotella2014state,hartley2020contact}. This contact-aided formulation is simple, interpretable, and real-time: when a foot is stationary with respect to the ground, the kinematic chain provides a strong constraint on body motion.

\begin{figure}[!t]
\centering
\includegraphics[width=\columnwidth]{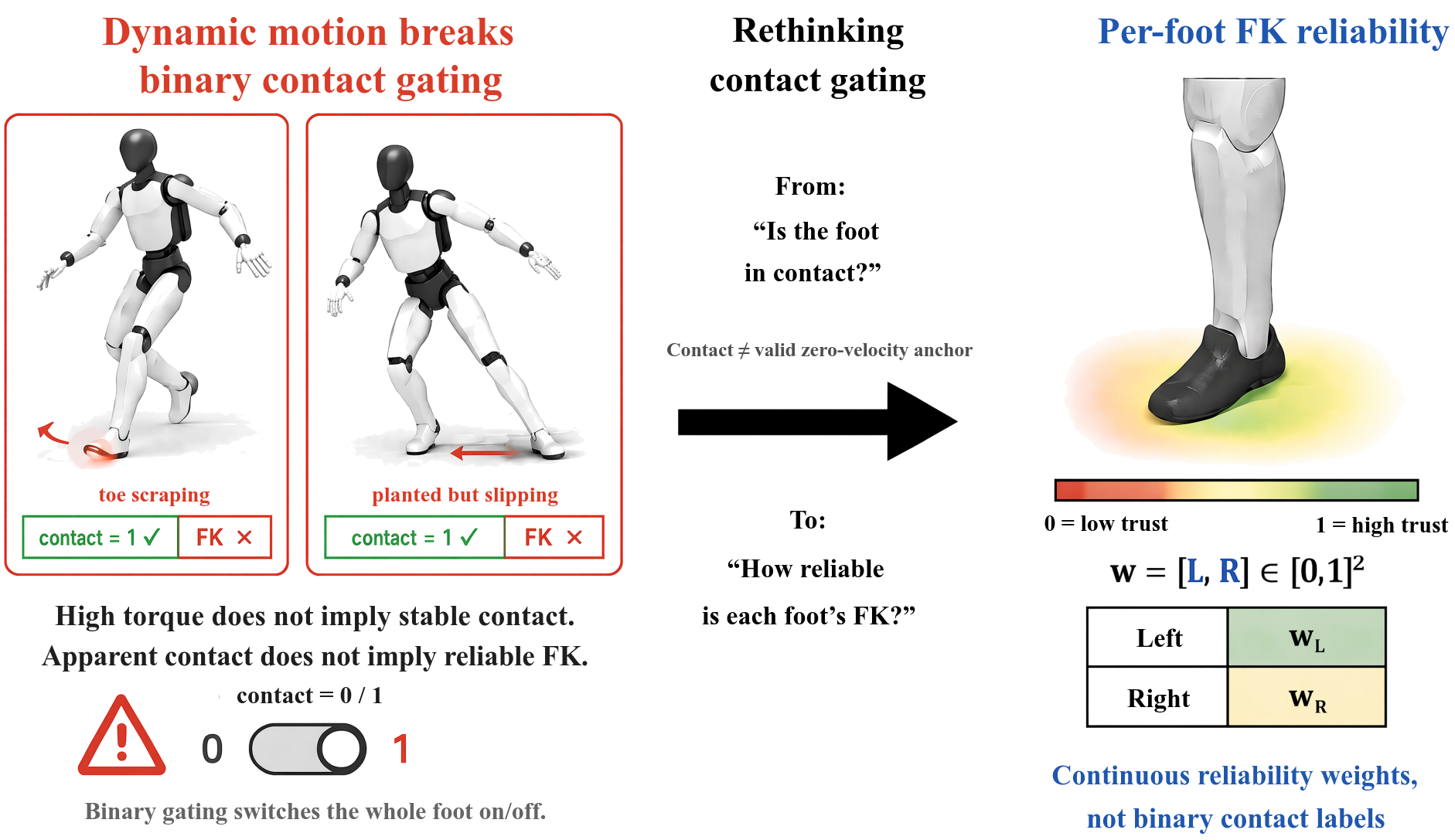}
\caption{\textbf{Motivation of FOCUS.} Dynamic humanoid motion breaks binary contact gating: high torque or apparent contact does not necessarily imply reliable FK. FOCUS instead estimates continuous per-foot FK reliability weights for the left and right feet.}
\label{fig:motivation}
\end{figure}

The difficulty is deciding when FK should be trusted. Existing systems typically infer binary contact from force, estimated torque, or learned classifiers and then gate the entire FK observation~\cite{hwangbo2016probabilistic,lin2021legged,kim2021legged}. However, long walking sequences amplify contact errors, while dynamic motions introduce toe scraping, partial support, fast swings, and slip. Because FK requires a rigid, stationary-foot constraint, a contacting foot may still provide an unreliable observation, which binary gating cannot express.

FOCUS (Foot Observation Confidence from Unannotated Simulation), illustrated in Fig.~\ref{fig:motivation}, reformulates contact handling as continuous FK reliability estimation. It predicts continuous per-foot weights that blend FK velocity observations with IMU-propagated body velocity and modulate the EKF observation covariance, while preserving the model-based FK-EKF structure.

The reliability estimator is a lightweight causal Transformer that uses a short history of IMU, joint position, and joint velocity measurements. It is trained from automatically generated simulation signals without manually annotated continuous reliability labels. The deployed model is sensor-only and does not use torque input, reducing sensitivity to actuator-current-based torque estimates.

Experiments show that FOCUS reduces mean ATE by 83.7\% in simulated walking, by 70.8\% across 19 real walking segments, and by 42.7\% across four real dynamic-motion routines.

\begin{enumerate}
    \item We construct a simulation data-collection pipeline for FK reliability learning. A humanoid motion-tracking control policy is trained in Isaac Lab and then replayed over diverse motion clips, producing executed robot motions with domain randomization, tracking error, and contact variation for training.

    \item We train a causal Transformer to estimate two continuous per-foot FK reliability weights in $[0,1]$ from sensor-only proprioception. The continuous reliability output reduces sim-to-real brittleness compared with hard contact labels, and the FK-weighted velocity consistency objective removes the need for manually annotated continuous FK-reliability labels.

    \item We evaluate the learned estimator on simulation, a 19-segment real walking set collected from five A3 Ultra humanoid units, and 18 real dynamic-motion sequences across four dance types, against threshold-based and representative proprioceptive odometry baselines.
\end{enumerate}

\section{Related Work}
\label{sec:related_work}

\subsection{Threshold-Based Contact-Aided Odometry}

Classical contact-aided odometry uses the fact that a stance foot provides a kinematic constraint on body motion. EKF and invariant EKF formulations fuse IMU measurements with leg kinematics under this assumption~\cite{bloesch2013state,hartley2020contact,camurri2020pronto}. In practical systems, the foot constraint is usually enabled or disabled by hand-designed thresholds on force, estimated torque, or contact-state logic~\cite{hwangbo2016probabilistic,kim2021legged}. Recent proprioceptive and multi-sensor odometry systems further combine leg kinematics with vision, light detection and ranging (LiDAR), or additional IMUs~\cite{wisth2022vilens,yang2023cerberus,ou2024legkilo,yang2023multiimu,shan2021lvisam,xu2022fast,zheng2025fastlivo2}, showing that leg kinematics remains useful even in richer sensor pipelines.

The common bottleneck is contact quality: FK is useful only when the foot-ground constraint is valid. Threshold-based gating is interpretable and efficient, but it makes a hard whole-foot decision and is coarse for toe contact, slip, or partial support. FOCUS targets this bottleneck directly by replacing binary threshold gating with continuous FK reliability.

\subsection{Learned Contact Estimation}

To reduce the brittleness of hand thresholds, learning has been used to estimate contact-related quantities from proprioception. Lin et al.~\cite{lin2021legged} used learned contact events with an invariant Kalman filter. Youm et al.~\cite{youm2025neural} integrated a neural measurement network (NMM) with an invariant EKF to predict measurement quantities such as contact probability and base velocity. Concurrent estimator--controller training and Transformer-based state estimation further show the benefit of temporal proprioceptive models for agile locomotion~\cite{ji2022concurrent,yu2024set}. More recently, CoCo-InEKF~\cite{baumgartner2026coco} learned continuous contact covariances for an invariant EKF, avoiding hand-designed binary contact thresholds in contact-rich locomotion.

These approaches differ in their learning target, training signal, and coupling to the filter. The NMM predicts a replacement measurement, such as base velocity, together with contact-related outputs, and is optimized through an EKF state-estimation objective. Its predicted measurement is inserted into the filter rather than used to assess an existing FK measurement. CoCo-InEKF predicts a per-foot contact covariance and adapts the filter covariance, while leaving the FK measurement itself unchanged. Neither formulation explicitly trains a foot-specific reliability signal to select between FK-derived body-velocity observations according to their consistency.

FOCUS learns a per-foot FK-velocity reliability weight tied to the FK-derived body-velocity observation used by the EKF. Its primary training signal is the FK-weighted velocity-consistency loss against simulated body velocity, with only lightweight binary-contact BCE regularization; it does not require manually annotated continuous reliability or covariance labels. At deployment, the weight changes the confidence assigned to FK and blends the FK velocity observation with the IMU-propagated velocity before the EKF update. Contact probability, learned contact covariance, learned replacement measurement, and FK reliability are distinct quantities: a foot can be in contact while toe scraping, partial support, or slipping makes its FK velocity unreliable.

\subsection{End-to-End Proprioceptive Odometry}

Another line of work bypasses explicit contact modeling and learns odometry or displacement directly from proprioceptive signals. Buchanan et al.~\cite{buchanan2022learning} learned inertial odometry for dynamic legged state estimation, and Legolas~\cite{wasserman2025legolas} learned deep leg-inertial odometry from robot proprioception. Related recent work also learns leg-kinematic motion factors or uncertainty inside larger factor-graph systems~\cite{okawara2025tightly}. These end-to-end approaches can capture complex contact effects, but the learned module often carries more of the metric odometry burden.

Our approach keeps the metric FK computation and EKF update explicit. The network only estimates how much each foot should be trusted. This bounded reliability output is easier to interpret than direct displacement prediction and can be integrated conservatively by blending FK velocity observations and inflating measurement noise. The result is a middle ground between threshold-based contact gating, learned contact estimation, and pure end-to-end proprioceptive odometry: FOCUS retains a real-time model-based estimator while learning the dynamic reliability of its FK observations.

\section{Methodology}
\label{sec:method}

\subsection{System Overview}

\begin{figure*}[!t]
\centering
\includegraphics[width=\textwidth]{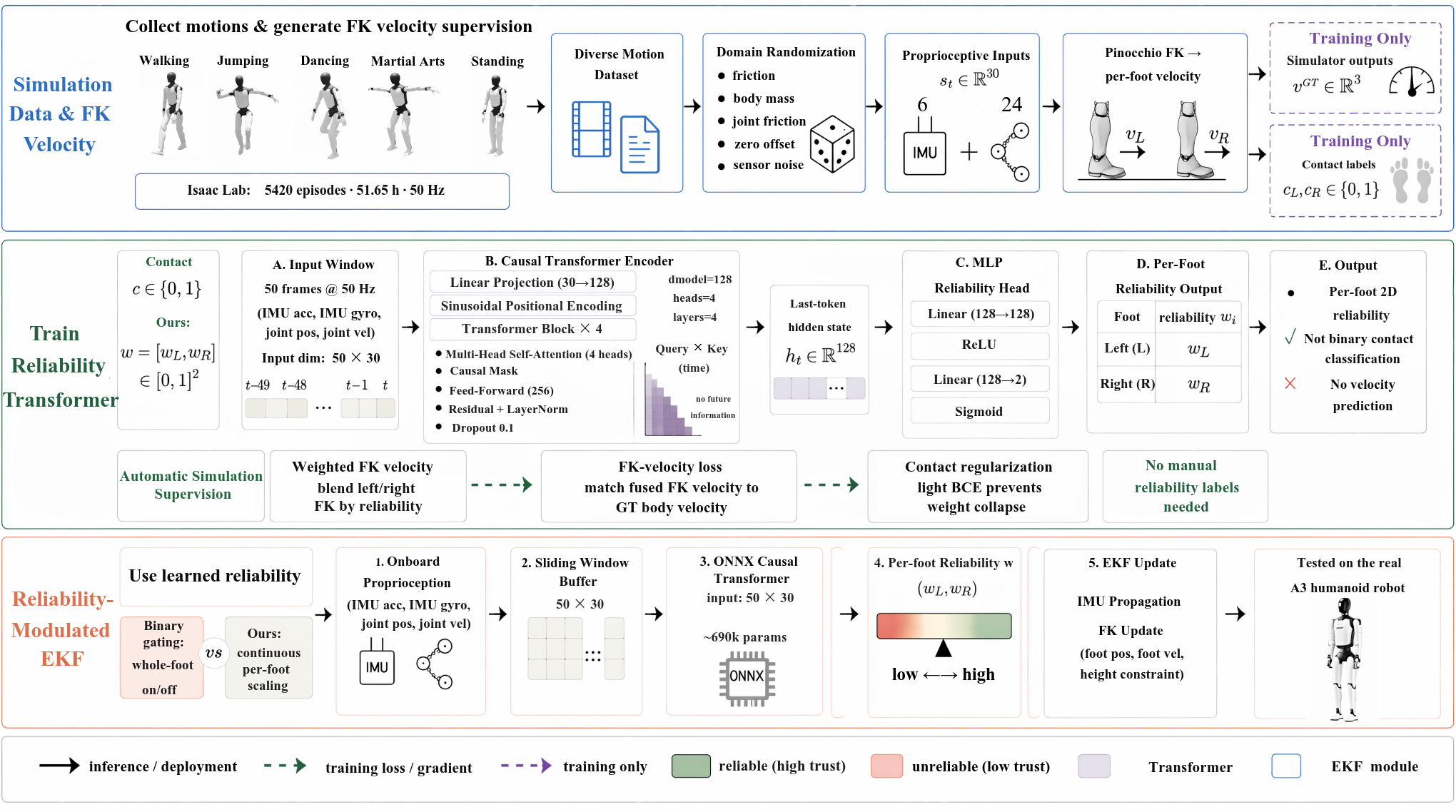}
\caption{\textbf{Per-foot FK Reliability Weighting.} A causal Transformer predicts left/right FK reliability weights from a 50-frame proprioceptive history. The weights blend FK and IMU-propagated velocity observations and modulate EKF noise; training uses FK-weighted velocity consistency with lightweight contact regularization.}
\label{fig:method_overview}
\end{figure*}

We combine a model-based EKF with a learned FK reliability estimator, as shown in Fig.~\ref{fig:method_overview}. The EKF maintains body position, body velocity, and foot positions using IMU propagation and FK-derived foot observations. The learned module does not output odometry states. It outputs two reliability weights that determine how much the EKF should trust the left and right foot FK observations.

At each estimator update, joint positions and velocities are used to compute foot positions and velocities through Pinocchio forward kinematics~\cite{carpentier2019pinocchio}. A causal Transformer processes a short history of proprioceptive measurements and predicts two reliability weights $w_L,w_R\in[0,1]$. The EKF uses these weights to blend the FK velocity observation with the IMU-propagated body velocity and inflate FK observation noise.

\subsection{EKF State and Observation Model}
\label{sec:ekf}

The filter uses a 12-dimensional state:
\begin{equation}
\label{eq:state_vector}
\x =
\begin{bmatrix}
\left(\mathbf{p}_B^G\right)^{\top} &
\left(\mathbf{v}_B^G\right)^{\top} &
\left(\mathbf{p}_{f_L}^G\right)^{\top} &
\left(\mathbf{p}_{f_R}^G\right)^{\top}
\end{bmatrix}^{\top}\in\R^{12}
\end{equation}
where body and foot positions and body velocity are expressed in the global frame. We use $\mathbf{p}_C^A$ and $\mathbf{v}_C^A$ for quantities of $C$ expressed in $A$, and $\mathbf{R}_{AB}$ maps vectors from $B$ to $A$. Orientation is not part of the 12-dimensional translational state: at each update, $\mathbf{R}_{GB}$ is obtained directly from the normalized pelvis-IMU quaternion and treated as a measured input, while $\bs{\omega}_B^B$ is obtained from the gyroscope. The body position and velocity are propagated with gravity-compensated IMU acceleration; the filter does not estimate attitude or IMU biases, and foot-state process noise increases for unreliable feet.

The 14-dimensional observation used in all experiments comprises body-relative FK positions and FK-derived body velocities for both feet, together with two scalar foot-height constraints. For foot $i\in\{L,R\}$, Pinocchio provides the body-frame position $\mathbf{p}_{f_i,\mathrm{FK}}^B$ and Jacobian velocity $\mathbf{v}_{f_i,\mathrm{FK}}^B$. Under the stationary-foot constraint, the resulting body-velocity observation is
\begin{equation}
\label{eq:fk_velocity}
\mathbf{v}_{B,i}^{G,\mathrm{FK}}
=
-\mathbf{R}_{GB}
\left(
\bs{\omega}_{B}^B\times\mathbf{p}_{f_i,\mathrm{FK}}^B
+\mathbf{v}_{f_i,\mathrm{FK}}^B
\right).
\end{equation}
This observation requires a valid non-slipping ground constraint, whose reliability is estimated continuously by FOCUS.

\subsection{Per-Foot FK Reliability Modulation}
\label{sec:reliability_modulation}

For foot $i\in\{L,R\}$, FOCUS replaces binary contact gating with continuous modulation. The FK velocity observation covariance is
\begin{equation}
\label{eq:R_modulation}
R_{\mathrm{vel},i} = R_{\mathrm{vel},0}\left[1 + (1-w_i)S_{\mathrm{vel}}\right]
\end{equation}
where $R_{\mathrm{vel},0}$ is nominal FK velocity noise and $S_{\mathrm{vel}}$ is the maximum amplification. The same modulation is applied to foot-position covariance, foot-height covariance, and foot-state process noise:
\begin{equation}
\label{eq:other_noise_modulation}
\left\{
\begin{aligned}
R_{p,i} &= R_{p,0}\left[1+(1-w_i)S_p\right],\\
R_{h,i} &= R_{h,0}\left[1+(1-w_i)S_h\right],\\
Q_{f,i} &= Q_{f,0}\left[1+(1-w_i)S_q\right],
\end{aligned}
\right.
\end{equation}

For all reported experiments, $S_{\mathrm{vel}}=100$, $S_p=1000$, and $S_h=S_q=100$. The corresponding nominal diagonal noise blocks are $R_{\mathrm{vel},0}=0.01\mathbf{I}_3$, $R_{p,0}=0.1\mathbf{I}_3$, $R_{h,0}=0.1$, and $Q_{f,0}=0.002\Delta t\mathbf{I}_3$ for each foot.

The weight also modulates the velocity observation. A saturated ramp $\phi(\cdot)$ maps $w_i$ to the trust coefficient $\tau_i=\phi(w_i)\in[0,1]$:
\begin{equation}
\label{eq:trust_ramp}
\tau_i=\phi(w_i)=\mathrm{clip}\left(\frac{w_i}{w_{\mathrm{sat}}},0,1\right),
\quad w_{\mathrm{sat}}=0.4
\end{equation}
which is equivalent to $\tau_i=\min(1,2.5w_i)$ because $w_i\in[0,1]$.
The velocity observation supplied to the EKF is
\begin{equation}
\label{eq:velocity_blending}
\z_{v,i} =
(1-\tau_i)\,\hat{\mathbf{v}}_{B,k|k-1}^G
+ \tau_i\,\mathbf{v}_{B,i}^{G,\mathrm{FK}}
+ \mathbf{n}_{v,i}
\end{equation}
Here $\hat{\mathbf{v}}_{B,k|k-1}^G$ is the pre-update IMU velocity and $\mathbf{v}_{B,i}^{G,\mathrm{FK}}$ is from Eq.~\ref{eq:fk_velocity}. Low reliability favors IMU propagation, whereas high reliability relies on FK; raw $w_i$ controls covariance scaling and $\phi(w_i)$ controls velocity trust.

\subsection{Causal Transformer Reliability Estimator}
\label{sec:transformer}

\subsubsection{Input Features}

The reliability network uses a 30-dimensional sensor-only proprioceptive input at each frame:
\begin{equation}
\mathbf{s}_t = \begin{bmatrix}
\mathbf{a}_t^\top & \bs{\omega}_t^\top & \mathbf{q}_t^\top & \dot{\mathbf{q}}_t^\top
\end{bmatrix}^\top \in \R^{30}
\end{equation}
where $\mathbf{a}_t\in\R^3$ is IMU linear acceleration, $\bs{\omega}_t\in\R^3$ is IMU angular velocity, and $\mathbf{q}_t\in\R^{12}$ and $\dot{\mathbf{q}}_t\in\R^{12}$ are lower-limb joint positions and velocities. The final model removes joint torque input because estimated torque is sensitive to current-to-torque calibration and hardware bias. The model processes a temporal window of $T=50$ frames, corresponding to one second at the 50\,Hz training frequency.

All channels are normalized using training-set mean and standard deviation. The normalized 30-dimensional sequence is fed directly to the reliability network without additional finite-difference augmentation.

\subsubsection{Layer Structure}

The model projects each frame to a 128-dimensional token, adds fixed sinusoidal positional encoding, and processes the sequence with four causal Transformer encoder layers~\cite{vaswani2017attention} using 4-head self-attention. A two-layer reliability head matching the deployed Open Neural Network Exchange (ONNX) model maps the final token to two sigmoid weights $[w_L,w_R]$.

\subsection{FK-Weighted Training Objective}
\label{sec:loss}

Continuous FK reliability is intrinsically ambiguous and difficult to annotate. We therefore train the network using automatically generated simulation signals, without manually annotated continuous FK-reliability labels, based on the principle that reliable FK velocity should agree with the ground-truth (GT) body velocity. Let $\mathbf{v}^{\mathrm{FK}}_{L}$ and $\mathbf{v}^{\mathrm{FK}}_{R}$ be the body velocity vectors inferred from the left and right feet. The network weights these two estimates as
\begin{equation}
\hat{\mathbf{v}} = \frac{w_L\mathbf{v}^{\mathrm{FK}}_{L}+w_R\mathbf{v}^{\mathrm{FK}}_{R}}{w_L+w_R+\epsilon}
\end{equation}
where $\epsilon$ avoids division by zero. The primary loss is
\begin{equation}
\label{eq:loss_fk}
\mathcal{L}_{\mathrm{FK}} = \left\|\hat{\mathbf{v}}-\mathbf{v}^{\mathrm{GT}}\right\|_2^2\left(1+5\|\mathbf{v}^{\mathrm{GT}}\|_1\right).
\end{equation}
The speed-dependent factor emphasizes high-speed motions, where incorrect FK reliability produces larger odometry drift.

We also use binary cross-entropy (BCE) for lightweight contact regularization:
\begin{equation}
\mathcal{L}_{\mathrm{BCE}} = \mathrm{BCE}(w_L,c_L)+\mathrm{BCE}(w_R,c_R)
\end{equation}
with
\begin{equation}
\mathrm{BCE}(w,c)=-c\log w-(1-c)\log(1-w)
\end{equation}
where $c_L,c_R\in\{0,1\}$ are simulator contact states. The total loss is
\begin{equation}
\mathcal{L} = \mathcal{L}_{\mathrm{FK}} + 0.3\mathcal{L}_{\mathrm{BCE}}
\end{equation}
The FK-weighted velocity loss provides the main task supervision, while the simulator-contact BCE term serves only as an auxiliary regularizer that prevents degenerate weights. Neither term requires manually annotated continuous FK-reliability labels.

\subsection{Simulation Data Collection}
\label{sec:data_collection}

\begin{figure}[!t]
\centering
\includegraphics[width=\columnwidth]{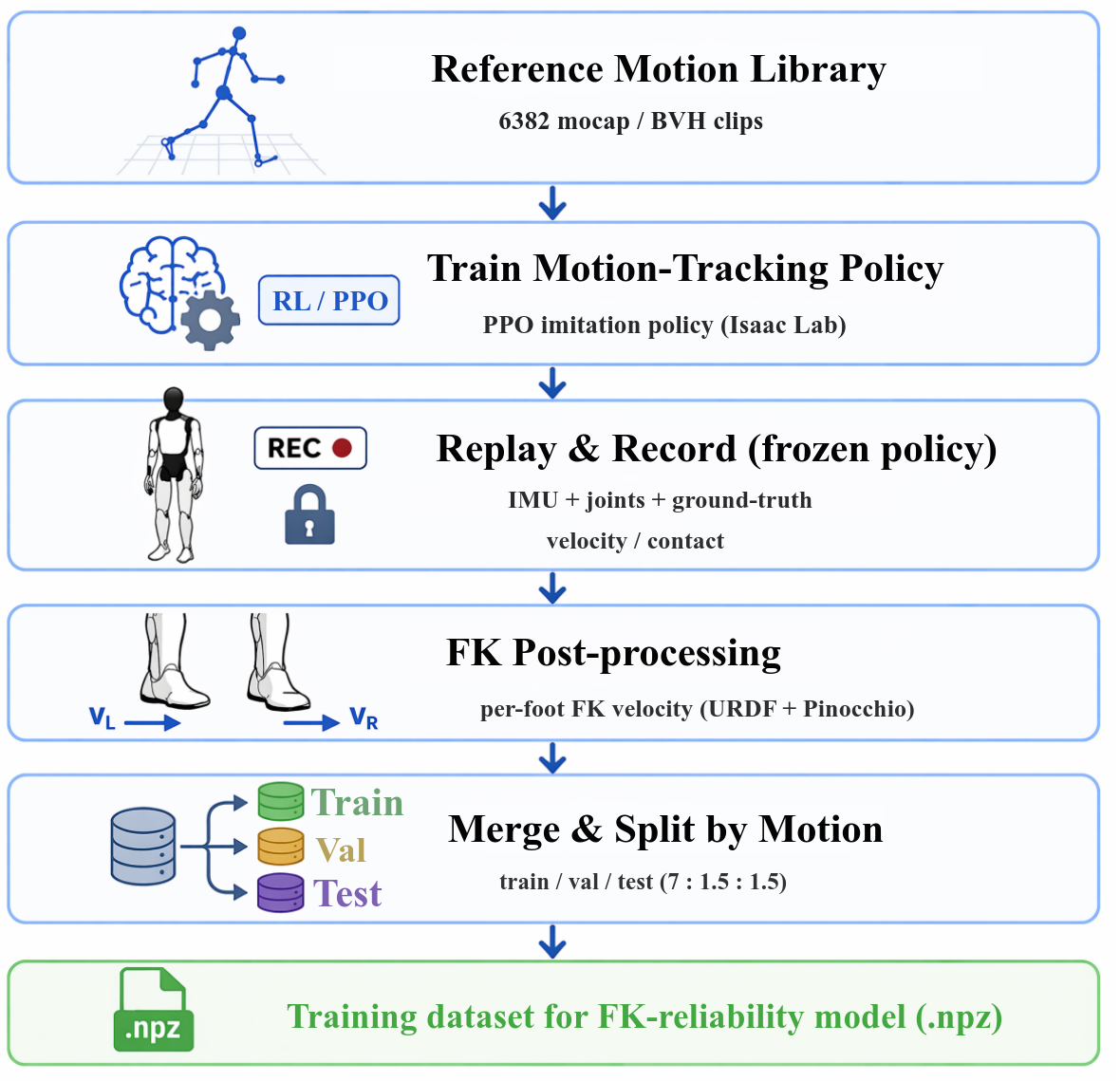}
\caption{Isaac Lab simulation data collection pipeline. A reinforcement-learning (RL) policy trained with proximal policy optimization (PPO) tracks retargeted motions, is frozen for replay-based recording, and is post-processed with a Unified Robot Description Format (URDF) model to produce per-foot FK velocity supervision before motion-wise dataset splitting.}
\label{fig:sim_data_pipeline}
\end{figure}

The motion library combines clips from open-source motion datasets with motions captured by a real motion-capture (MoCap) system. We collect training data with the Isaac Lab/Orbit motion-tracking pipeline~\cite{mittal2023orbit} in Fig.~\ref{fig:sim_data_pipeline}, rather than with random velocity commands. Following recent humanoid motion-tracking work~\cite{luo2025sonic}, a proximal policy optimization (PPO) controller tracks retargeted clips. We freeze the policy, replay it over the combined motion library, and record executed states instead of reference motions to retain tracking errors, imperfect contacts, actuator effects, and randomized dynamics.

Four domain-randomized repeats cover walking, jumping, standing, running, and martial-arts motions. The splits contain 5420 training episodes and 1151 validation episodes; training spans about 51.65\,h at 50\,Hz. The recorder stores IMU, body-orientation, joint-state, torque, and motion-phase signals, together with body-velocity, root-pose, and contact labels; FOCUS uses only the IMU and joint channels in Sec.~\ref{sec:transformer}. Randomization covers friction, mass, joint properties, pose offsets, torso center of mass, external pushes, Gaussian sensor noise, and episode bias, while labels remain clean.

We generate FK velocity supervision offline after recording. The smoothed joint states are processed with the robot URDF model and Pinocchio; the two foot Jacobians provide body-velocity estimates according to Eq.~\ref{eq:fk_velocity}. These velocities are training signals only. Batches are split by motion identity so that repeated executions of one reference motion do not cross partitions.

\section{Experiments}
\label{sec:experiments}

We evaluate FOCUS on simulated walking and dynamic motions, 19 real walking segments, and real dynamic-motion sequences covering four routines: Charleston, Chaosha dance, Zero-Frame Start, and Cyberwalk. The comparisons include the original torque-threshold contact-gating baseline and three external proprioceptive odometry methods.

\subsection{Experimental Setup}

\begin{figure}[!t]
\centering
\includegraphics[width=0.96\columnwidth]{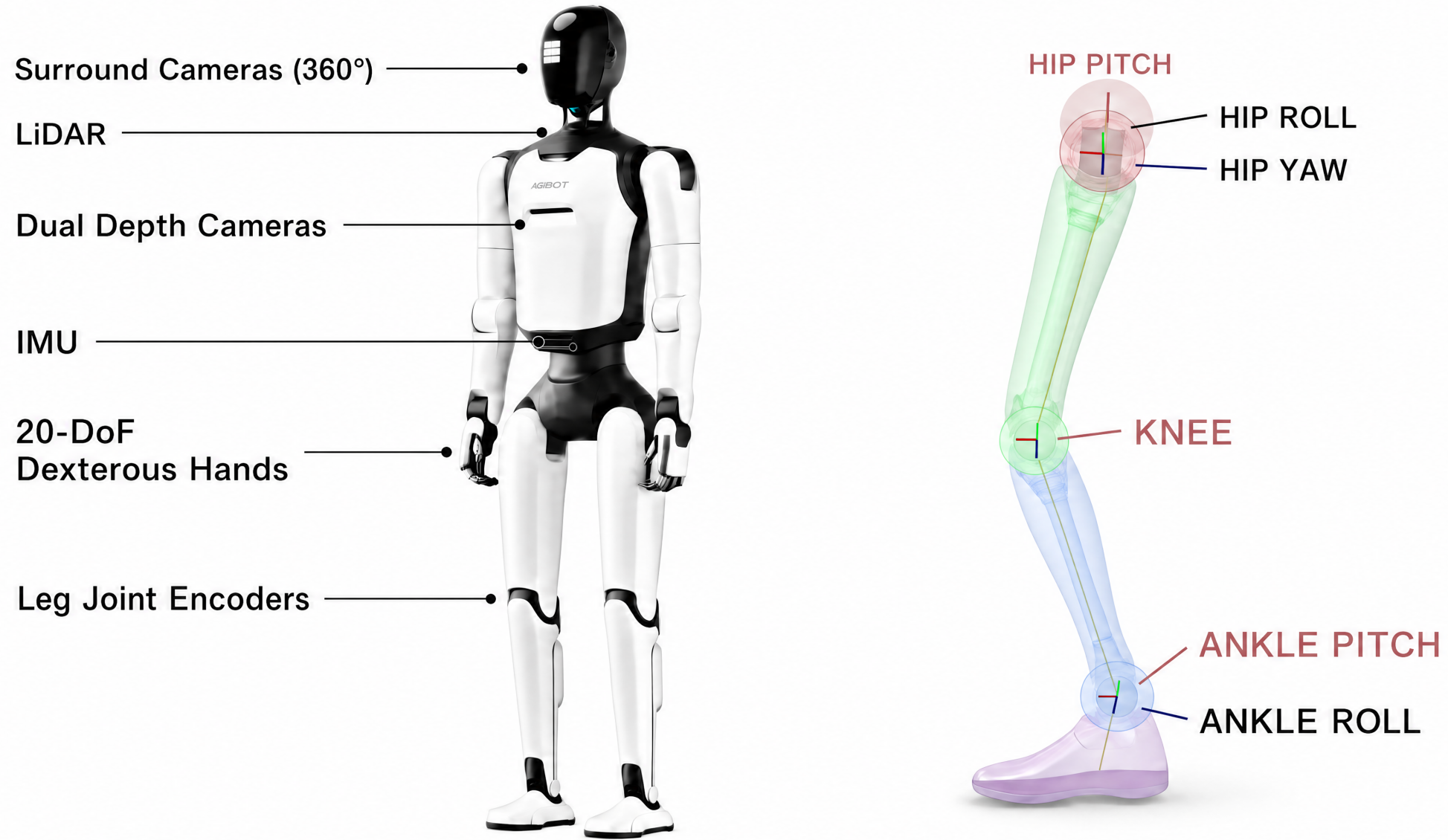}
\caption{A3 Ultra humanoid platform used for hardware evaluation. The left panel summarizes onboard sensors, including LiDAR, surround-view cameras, dual depth cameras, an IMU, and leg joint encoders; the right panel shows the lower-limb URDF kinematic structure used for FK. FOCUS uses only the proprioceptive IMU and lower-limb joint-encoder measurements, while exteroceptive sensors are not used by the odometry estimator.}
\label{fig:a3_platform}
\end{figure}

Hardware experiments use the A3 Ultra humanoid platform shown in Fig.~\ref{fig:a3_platform}, a full-size robot with 174\,cm height and 60\,kg mass. It carries LiDAR, surround cameras, dual depth cameras, a pelvis IMU, lower-limb encoders, and a calibrated leg model for FK odometry. Although exteroceptive sensors are available, FOCUS uses only proprioception. The data come from five A3 Ultra units sharing the same kinematic model and sensor interface. The deployed model uses the 30-dimensional input described in Section~\ref{sec:transformer}.

Trajectory errors are reported on the horizontal XY plane after timestamp matching. Simulation ground truth is obtained directly from the simulator. For the real-robot experiments, a LiDAR-based reference trajectory serves as ground truth for walking, whereas a motion-capture (MoCap) system provides ground truth for dynamic motions; both are used only for offline evaluation and are not supplied to any estimator. Simulated walking uses XY start-point alignment; simulated dynamic motions and real dynamic motions use normalized-time interpolation with best-fit yaw alignment; real walking uses per-segment yaw-and-translation alignment. No scale correction is applied. For timestamp-wise horizontal errors $e_t$, ATE is the root-mean-square error $\sqrt{N^{-1}\sum_{t=1}^{N}e_t^2}$; MAE, MED, and STD are respectively the mean, median, and standard deviation of $e_t$, and Drift is the ATE normalized by ground-truth path length. Results spanning multiple sequences are reported as arithmetic means of their per-sequence metrics.

We compare FOCUS with the original torque-threshold estimator, which uses estimated joint torque to gate whole-foot FK observations, and with Pronto~\cite{camurri2020pronto}, CoCo-InEKF~\cite{baumgartner2026coco}, and Legolas~\cite{wasserman2025legolas}. All methods use the same A3 Ultra URDF, Pinocchio FK implementation, ankle-roll foot frame, timestamp synchronization, and no external localization input, while retaining their original estimator formulations and proprioceptive inputs. When original checkpoints were incompatible, learning-based baselines were retrained or adapted only on the Isaac Lab training split; no real evaluation sequence was used. Legolas was originally designed for quadrupeds, creating a morphology and motion-domain mismatch on A3 Ultra. FOCUS is the per-foot sensor-only model trained exclusively on the simulation data in Section~\ref{sec:data_collection}. The real benchmark contains 19 walking segments from five robots (1.51\,km, 54.1\,min) and dynamic sequences from four routines.

\begin{figure*}[!t]
\centering
\includegraphics[width=\textwidth]{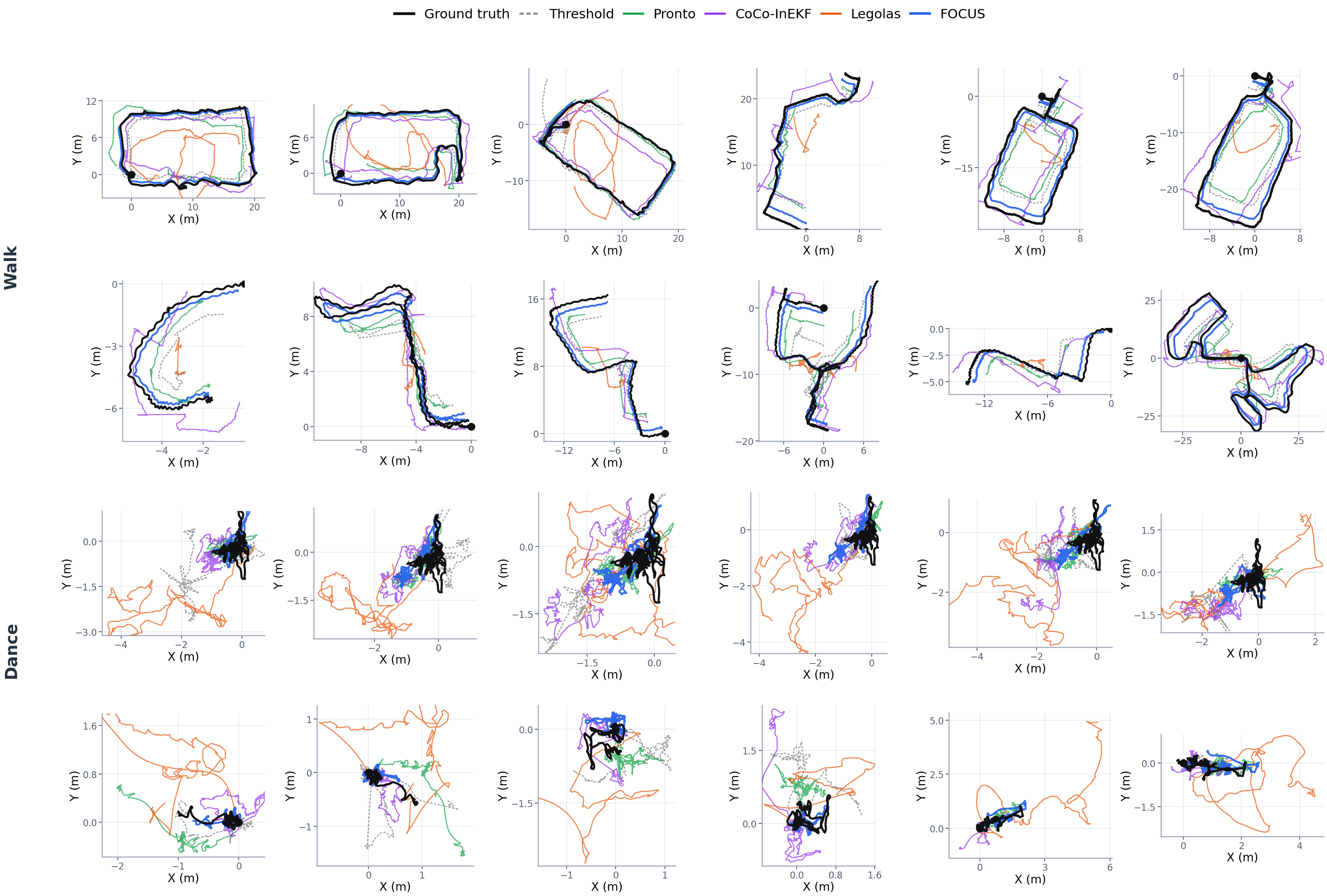}
\caption{Real-robot trajectory comparisons across representative walking and dynamic-motion sequences. The first two rows show walking examples selected from the 19-segment set. The last two rows show dynamic motions spanning Charleston, Chaosha dance, Zero-Frame Start, and Cyberwalk. The plot compares ground truth, Threshold, Pronto, CoCo-InEKF, Legolas, and FOCUS.}
\label{fig:real_all_trajectories}
\end{figure*}

\subsection{Simulation Evaluation}
\label{sec:sim_eval}

The simulation evaluation uses 20 Isaac Lab evaluation episodes each for walking and dynamic motions, exposing accumulated contact errors across varied motion patterns.

\begin{table}[!t]
\centering
\caption{Simulated walking comparison over 20 episodes. Errors are in meters except Drift (\%); lower is better.}
\label{tab:sim_walk_results}
\footnotesize
\setlength{\tabcolsep}{2.8pt}
\renewcommand{\arraystretch}{1.05}
\begin{tabular}{lccccc}
\toprule
Method & ATE & MAE & MED & STD & Drift (\%) \\
\midrule
Threshold & 1.016 & 0.950 & 1.067 & 0.363 & 5.47 \\
Pronto~\cite{camurri2020pronto} & 0.517 & 0.469 & 0.452 & 0.214 & 1.59 \\
CoCo-InEKF~\cite{baumgartner2026coco} & 1.283 & 1.190 & 1.255 & 0.479 & 9.13 \\
Legolas~\cite{wasserman2025legolas} & 1.324 & 1.208 & 1.314 & 0.539 & 8.55 \\
\textbf{FOCUS} & \textbf{0.166} & \textbf{0.156} & \textbf{0.168} & \textbf{0.056} & \textbf{0.73} \\
\bottomrule
\end{tabular}
\end{table}

On simulated walking, the proposed weighting gives the lowest ATE and drift, reducing mean ATE from 1.016\,m to 0.166\,m relative to torque-threshold gating (Table~\ref{tab:sim_walk_results}). Continuous FK reliability limits drift from persistent contact errors.

Dynamic motions can have small net displacement, so ATE alone may understate trajectory fidelity. We report amplitude fidelity, the estimated-to-ground-truth XY motion-extent ratio (with 1 indicating perfect agreement, below 1 indicating shrinkage, and above 1 indicating over-amplification); discrete Fr\'echet distance, a curve-shape error in meters (lower is better); and spectral ratio, the 0.5--3\,Hz motion-band position-energy ratio, with 1 indicating perfect agreement.

\begin{table}[!t]
\centering
\caption{Simulated dynamic-motion results over 20 episodes. ATE and Fr\'echet distance are in meters; for the Amp. and Spec. ratios, 1 indicates perfect agreement.}
\label{tab:sim_dynamic_results}
\footnotesize
\setlength{\tabcolsep}{2.8pt}
\begin{tabular}{lcccc}
\toprule
Method & ATE $\downarrow$ & Amp. ($\to 1$) & Fr\'echet $\downarrow$ & Spec. ($\to 1$) \\
\midrule
Threshold & \textbf{0.382} & 3.113 & \textbf{1.117} & 0.804 \\
Pronto~\cite{camurri2020pronto} & 0.932 & 1.294 & 1.620 & 0.673 \\
CoCo-InEKF~\cite{baumgartner2026coco} & 0.761 & 5.972 & 2.018 & 0.813 \\
Legolas~\cite{wasserman2025legolas} & 2.426 & 10.157 & 3.523 & 0.711 \\
\textbf{FOCUS} & 0.711 & \textbf{1.251} & 1.239 & \textbf{0.875} \\
\bottomrule
\end{tabular}
\end{table}

For the simulated dynamic set, FOCUS best preserves motion extent and spectral energy, while Threshold obtains the lowest ATE and Fr\'echet distance (Table~\ref{tab:sim_dynamic_results}). The complementary metrics are useful because these motions can have small net displacement and ATE alone may not reflect motion fidelity.

\subsection{Real Walking Evaluation}
\label{sec:real_walking}

Fig.~\ref{fig:real_all_trajectories} visualizes 12 representative segments from the 19-segment real-walking set.

All methods are evaluated on the same 19 segments (Table~\ref{tab:external_baselines}). Across this set, the proposed method reduces mean ATE from 2.634\,m to 0.768\,m, a 70.8\% reduction, and outperforms threshold gating on every segment.

\begin{table}[t]
\centering
\caption{Real-walking comparison on the same 19 segments. Errors are in meters except Drift (\%); lower is better.}
\label{tab:external_baselines}
\footnotesize
\setlength{\tabcolsep}{3.2pt}
\renewcommand{\arraystretch}{1.05}
\begin{tabular}{lccccc}
\toprule
Method & ATE & MAE & MED & STD & Drift (\%) \\
\midrule
Threshold & 2.634 & 2.371 & 2.281 & 1.100 & 3.85 \\
Pronto~\cite{camurri2020pronto} & 2.898 & 2.589 & 2.464 & 1.286 & 3.75 \\
CoCo-InEKF~\cite{baumgartner2026coco} & 3.290 & 2.931 & 2.896 & 1.477 & 4.47 \\
Legolas~\cite{wasserman2025legolas} & 8.235 & 7.466 & 7.020 & 3.405 & 9.64 \\
\textbf{FOCUS} & \textbf{0.768} & \textbf{0.695} & \textbf{0.677} & \textbf{0.319} & \textbf{0.92} \\
\bottomrule
\end{tabular}
\end{table}

On the paired 19-segment set, a two-sided Wilcoxon signed-rank test gives $p<10^{-5}$, indicating a consistent reduction across segments.

This transfer benefits from the sim-to-real design of FOCUS. Domain-randomized simulation covers friction, mass, joint properties, sensor noise, and disturbances, while the network outputs weights in $[0,1]$ rather than hard contact labels. The EKF can partially trust FK under ambiguous support, reducing sensitivity to threshold mismatch, actuator compliance, encoder noise, and torque-estimation error.

\subsection{Real Dance Evaluation}
\label{sec:dance_eval}

As a complementary dynamic-motion benchmark, we evaluate four routines on the A3 Ultra: Charleston, Chaosha dance, Zero-Frame Start, and Cyberwalk. The sequences are evaluated against motion-capture references and include non-periodic kicks, fast swings, partial support, toe contacts, and lateral motion, but little net displacement.

Table~\ref{tab:dance_fidelity} reports the aggregated metrics across the four routines. FOCUS achieves the lowest aggregate ATE among all compared methods, reducing mean ATE from 0.947\,m for Threshold to 0.542\,m; the next-best method, Pronto, reaches 0.605\,m. FOCUS also remains closest to perfect agreement in amplitude and spectral ratio while obtaining the lowest Fr\'echet distance. The expanded benchmark complements the long-displacement walking evaluation with a broader set of dynamic motions.

\begin{table}[!t]
\centering
\caption{Real dynamic-motion results across four routines: Charleston, Chaosha dance, Zero-Frame Start, and Cyberwalk. ATE and Fr\'echet distance are in meters; for the Amp. and Spec. ratios, 1 indicates perfect agreement.}
\label{tab:dance_fidelity}
\footnotesize
\setlength{\tabcolsep}{2.8pt}
\begin{tabular}{lcccc}
\toprule
Method & ATE $\downarrow$ & Amp. ($\to 1$) & Fr\'echet $\downarrow$ & Spec. ($\to 1$) \\
\midrule
Threshold & 0.947 & 1.537 & 1.744 & 3.804 \\
Pronto~\cite{camurri2020pronto} & 0.605 & 1.090 & 1.210 & 2.572 \\
CoCo-InEKF~\cite{baumgartner2026coco} & 0.847 & 1.312 & 1.450 & 2.430 \\
Legolas~\cite{wasserman2025legolas} & 1.956 & 2.229 & 3.148 & 7.888 \\
\textbf{FOCUS} & \textbf{0.542} & \textbf{0.953} & \textbf{0.777} & \textbf{1.327} \\
\bottomrule
\end{tabular}
\end{table}

\subsection{FK Reliability Versus Contact}
\label{sec:reliability_evidence}

Contact serves only as an indirect cue for FK quality. We therefore analyze the output using torque-independent kinematic signals over 139,013 frames of the real Charleston sequence. For each foot, the Spearman correlations between reliability and FK foot height and six-joint speed magnitude are $-0.608$ and $-0.663$, respectively, consistent with reduced FK reliability during foot lift and rapid motion.

\begin{figure}[!t]
\centering
\includegraphics[width=\columnwidth]{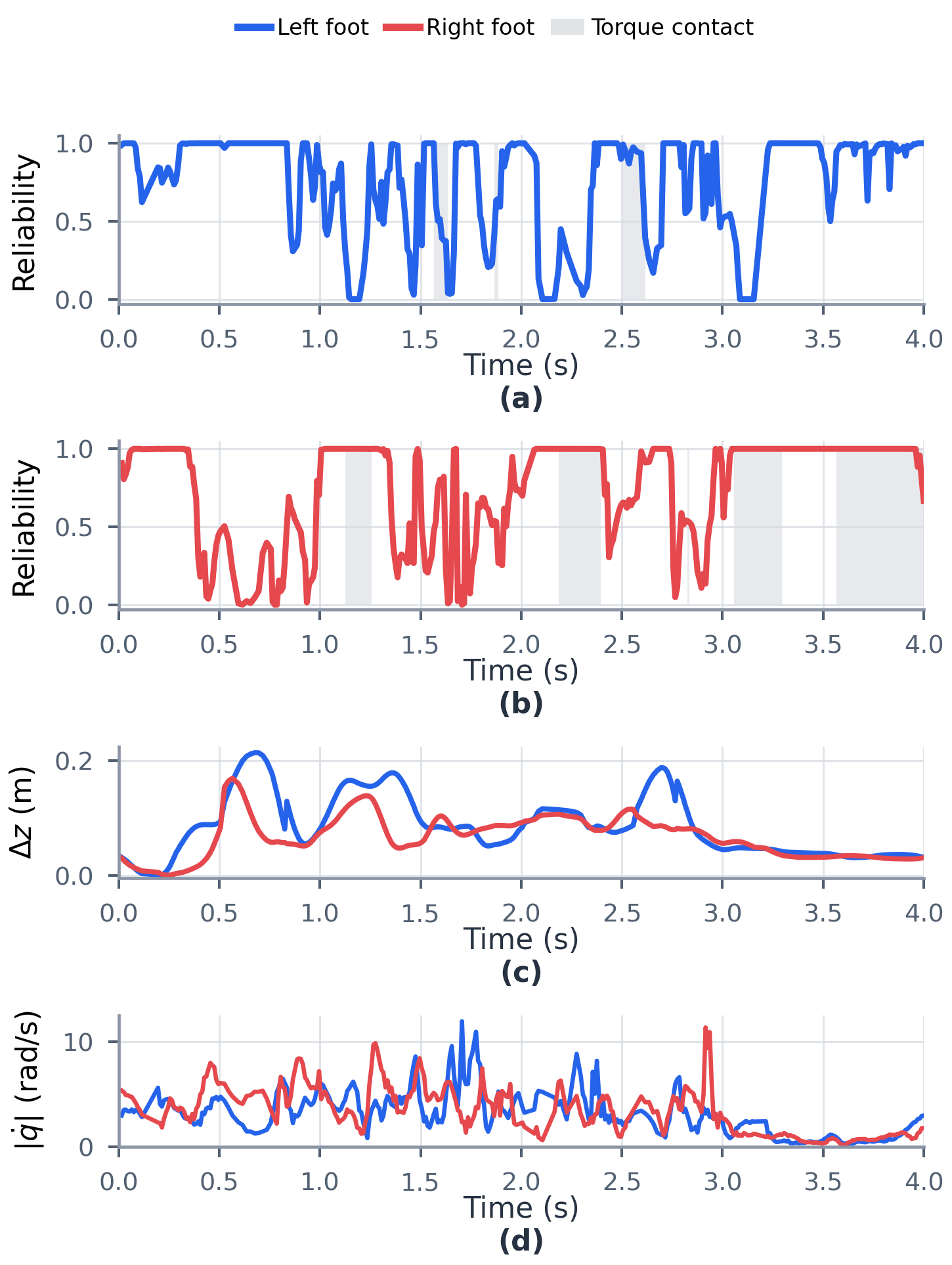}
\caption{Qualitative evidence that continuous FK reliability differs from binary contact in a representative 4-s Charleston window. Blue and red identify the left and right feet, respectively. Solid curves in (a) and (b) are the corresponding FOCUS reliability weights; gray shading denotes torque-based contact. Solid curves in (c) are relative FK foot heights $\Delta z$, while solid curves in (d) are leg-joint speeds.}
\label{fig:reliability_contact_evidence}
\end{figure}

Fig.~\ref{fig:reliability_contact_evidence} shows a representative 4-s interval containing contact transitions and intermediate reliability values. In (a), the blue solid curve is left-foot reliability; in (b), the red solid curve is right-foot reliability. Gray shading marks torque-based contact. Panel (c) shows the relative FK foot heights $\Delta z$ for both feet, while panel (d) shows leg-joint speed; $\Delta z$ is referenced to each foot's minimum height in the interval. The torque gate is a comparison cue, not a reliability label.

As described in Section~\ref{sec:reliability_modulation}, low reliability reduces FK velocity trust and increases the corresponding FK noise.

\subsection{Ablation Study}
\label{sec:ablation}

We compare five deployment choices on the same 19-segment real-walking set: hand thresholding, learned binary contact, thresholded continuous reliability, covariance-only modulation, and full FOCUS. Learned variants share the 30-dimensional input and training data. NN-binary outputs hard 0/1 contact; Hard FOCUS thresholds the learned weights at 0.5; FOCUS cov.-only disables Eq.~\ref{eq:velocity_blending} and is the closest ablation to learned contact-covariance filtering such as CoCo-InEKF~\cite{baumgartner2026coco}.

\begin{table}[!t]
\centering
\caption{Ablation on the same 19 real walking segments. Values are mean ATE in meters. Reductions are relative to torque-threshold contact gating.}
\label{tab:ablation_results}
\footnotesize
\setlength{\tabcolsep}{3.5pt}
\begin{tabular}{lccc}
\toprule
Method & Mode & ATE & Red. \\
\midrule
Threshold & hand bin. & 2.634 & -- \\
NN-binary & learned bin. & 6.125 & -132.5\% \\
Hard FOCUS & thresh. cont. & 0.802 & 69.5\% \\
FOCUS cov.-only & cov. only & 0.902 & 65.7\% \\
\textbf{FOCUS} & \textbf{cov.+blend} & \textbf{0.768} & \textbf{70.8\%} \\
\bottomrule
\end{tabular}
\end{table}

Continuous per-foot reliability is the strongest design (Table~\ref{tab:ablation_results}). NN-binary performs worse than hand thresholding, while hard-thresholding FOCUS and removing the velocity blend also degrade performance. The result supports both continuous modulation and the blend in Eq.~\ref{eq:velocity_blending}. On the paired set, two-sided Wilcoxon tests give $p<10^{-5}$ for Threshold versus FOCUS and covariance-only versus full FOCUS.

The 693,768-parameter ONNX model requires 1.62\,ms CPU time per inference on one Intel i7-13700K thread, using 8.1\% of one core at 50\,Hz and thus supporting real-time deployment.

\section{Conclusion}
\label{sec:conclusion}

FOCUS replaces binary contact gating with sensor-only, continuous per-foot FK reliability weights for EKF velocity blending and covariance modulation. It reduces ATE by 83.7\% in simulated walking, reduces mean ATE by 70.8\% across 19 real walking segments (from 2.634\,m to 0.768\,m), and reduces mean ATE by 42.7\% across four real dynamic-motion routines (from 0.947\,m to 0.542\,m).

Future work will extend FOCUS to additional motions and develop online adaptation to previously unseen terrain and contact conditions. During deployment, EKF innovations and short-horizon consistency between FK and IMU propagation could provide self-supervised update signals, while uncertainty-gated updates and regularization toward the pretrained model would help prevent unstable adaptation and catastrophic forgetting.


\bibliographystyle{IEEEtran}
\bibliography{references}

\end{document}